\documentclass[letterpaper]{article} % DO NOT CHANGE THIS
\usepackage{aaai25}  % DO NOT CHANGE THIS
\usepackage{times}  % DO NOT CHANGE THIS
\usepackage{helvet}  % DO NOT CHANGE THIS
\usepackage{courier}  % DO NOT CHANGE THIS
\usepackage[hyphens]{url}  % DO NOT CHANGE THIS
\usepackage{graphicx} % DO NOT CHANGE THIS
\usepackage{natbib}  % DO NOT CHANGE THIS AND DO NOT ADD ANY OPTIONS TO IT
\usepackage{caption} % DO NOT CHANGE THIS AND DO NOT ADD ANY OPTIONS TO IT
\usepackage{algorithm}
\usepackage{algorithmic}
\usepackage{amsmath,amsfonts}
\usepackage[capitalize]{cleveref}
\usepackage{cite}
\usepackage{multirow}
\usepackage{booktabs}
\usepackage{arydshln}
\usepackage{xcolor}
\newcommand{\etal}{et al.}
\usepackage{newfloat}
\usepackage{listings}
\DeclareCaptionStyle{ruled}{labelfont=normalfont,labelsep=colon,strut=off} % DO NOT CHANGE THIS
\floatstyle{ruled}
\newfloat{listing}{tb}{lst}{}
\floatname{listing}{Listing}
\title{Efficient Attention-Sharing Information Distillation Transformer\\ for Lightweight Single Image Super-Resolution}
\author {
    Karam Park\textsuperscript{\rm 1},
    Jae Woong Soh\textsuperscript{\rm 2\footnotemark[1]},
    Nam Ik Cho\textsuperscript{\rm 1,\rm 3\thanks{Corresponding authors}}
}
\affiliations {
    \textsuperscript{\rm 1}Department of ECE, INMC, Seoul National University, Seoul, Korea\\
    \textsuperscript{\rm 2}Department of EECS, Gwangju Institute of Science and Technology, Gwangju, Korea\\
    \textsuperscript{\rm 3}Interdisciplinary Program in Artificial Intelligence, Seoul National University, Seoul, Korea\\
    saturnian77@ispl.snu.ac.kr, jaewoongsoh@gist.ac.kr, nicho@snu.ac.kr
}

\begin{document}

\maketitle

\begin{abstract}
Transformer-based Super-Resolution (SR) methods have demonstrated superior performance compared to convolutional neural network (CNN)-based SR approaches due to their capability to capture long-range dependencies. However, their high computational complexity necessitates the development of lightweight approaches for practical use. To address this challenge, we propose the Attention-Sharing Information Distillation (ASID) network, a lightweight SR network that integrates attention-sharing and an information distillation structure specifically designed for Transformer-based SR methods. We modify the information distillation scheme, originally designed for efficient CNN operations, to reduce the computational load of stacked self-attention layers, effectively addressing the efficiency bottleneck. Additionally, we introduce attention-sharing across blocks to further minimize the computational cost of self-attention operations. By combining these strategies, ASID achieves competitive performance with existing SR methods while requiring only around 300$\,$K parameters – significantly fewer than existing CNN-based and Transformer-based SR models. Furthermore, ASID outperforms state-of-the-art SR methods when the number of parameters is matched, demonstrating its efficiency and effectiveness. The code and supplementary material are available on the project page.%Code and supplementary material are available at \url{https://github.com/saturnian77/ASID}.
\end{abstract}

\begin{links}
     %\link{Code $\&$ Supplementary Material}
     \link{Project Page}{https://github.com/saturnian77/ASID}
\end{links}

\section{Introduction}

Single Image Super-Resolution (SISR) is a technology designed to improve the resolution and quality of low-resolution (LR) images by transforming them into high-resolution (HR) counterparts. SISR is widely applied in tasks that require HR imagery, including medical imaging \cite{shi2013cardiac}, satellite imaging \cite{thornton2006sub}, and surveillance \cite{zhang2010super}. Despite its widespread use, SISR remains a challenging problem due to its inherent ill-posed nature.

Recently, SISR methods based on convolutional neural networks (CNNs) have shown significant performance improvement over traditional approaches \cite{timofte2013anchored,yang2010image,kim2010single}.
Researchers have progressively deepened the CNNs to improve performance using various techniques such as pixel-shuffle operation \cite{shi2016real} and residual connections \cite{he2016deep}. However, increasing network depth has resulted in a substantial rise in computational costs. Additionally, the introduction of attention mechanisms \cite{zhang2018image,dai2019second,mei2020image} has further escalated computational demands, limiting the practicality of SR models.
As a result, there is growing interest in developing efficient, lightweight networks that balance resource constraints and reconstruction performance.

To address the computational burden of CNNs, various strategies have been introduced, such as weight-sharing \cite{kim2016deeply,tai2017image,tai2017memnet,li2019feedback} and feature reuse \cite{he2019ode,luo2020latticenet,park2021single}. However, these approaches often impose limitations on network capabilities and may compromise performance due to restricted depth and receptive fields inherent in convolution operations.

\begin{figure}
  \centering
    \centering
    \includegraphics[width=\linewidth]{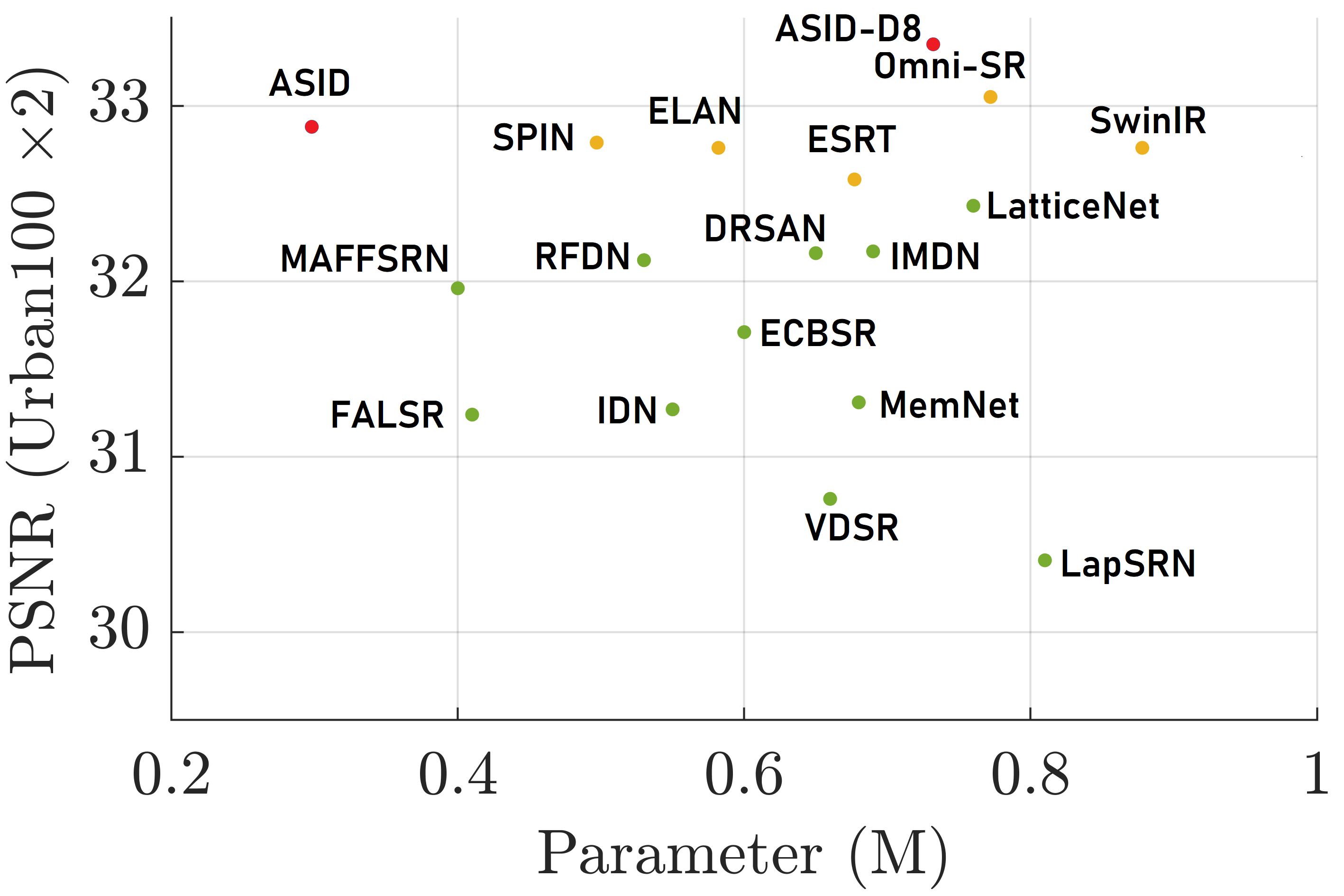}
  \caption{Visualized comparison of PSNR and the number of parameters on the Urban100 ($\times2$) dataset. Our ASID is compared with state-of-the-art lightweight SR methods. Green markers represent CNN-based methods, while yellow markers denote Transformer-based methods.}
  \label{fig:perfcomp}
\end{figure}

Recently, Transformers~\cite{vaswani2017attention} are increasingly being adopted for image reconstruction tasks, including SISR. Transformer-based SR methods have outperformed CNN-based methods by leveraging non-local context and globally expanding the receptive field. However, the computational intensity of self-attention layers in Transformers poses significant challenges for practical application in SISR. To address this, Liang {\etal}~\shortcite{liang2021swinir} proposed a patch-based processing technique to reduce computational demands, though this method limits the receptive field to the local window size. On the other hand, Zamir {\etal}~\shortcite{zamir2022restormer} introduced self-attention layers that utilize channel correlations to reduce computational costs, yet this approach fails to capture spatial correlations. While these methods represent notable progress in alleviating computational burdens, further advancements are required to improve reconstruction performance and efficiency.

\begin{figure}
  \centering
    \centering
    \includegraphics[width=\linewidth]{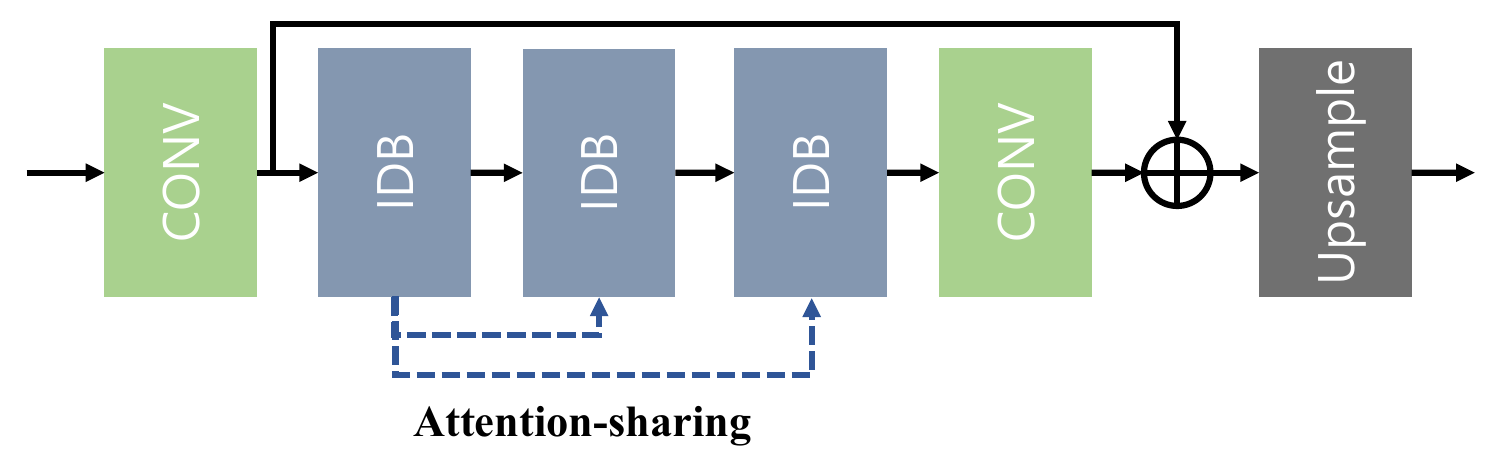}
  \caption{Visualized overall structure of ASID. ASID mainly consists of a convolutional layer for shallow feature extraction, a series of IDBs, and an upsampling module that reconstructs features into an SR image. Blue arrows indicate the attention-sharing mechanism.}
  \label{fig:overall}
\end{figure}

To address the challenges in designing efficient SR Transformers, we introduce a lightweight design scheme called the Attention-Sharing Information Distillation (ASID) network. Unlike previous approaches such as ESRT~\cite{lu2022transformer}, which adapted lightweight CNN frameworks to Transformers with limited success, ASID leverages an understanding of the distinct roles of convolution and self-attention layers~\cite{park2022vision} by integrating them into its architecture. By designing these layers to complement each other and adapting the information distillation scheme~\cite{hui2018fast,hui2019lightweight,liu2020residual} for Transformer architectures, ASID achieves state-of-the-art performance with remarkably few parameters. Additionally, we incorporate an attention-sharing method and eliminate correlation matrix calculations, significantly reducing computational costs without compromising reconstruction quality. With only about 300$\,$K parameters, ASID not only competes effectively with traditional CNN-based and Transformer-based SR methods but also surpasses state-of-the-art SISR networks of comparable complexities when scaled up.

Our contributions can be summarized in three-folds:
\begin{itemize}
    \item We introduce a new lightweight SR model that combines an information distillation design scheme with a Transformer framework.
    \item We propose a feature distillation framework with attention sharing to alleviate the efficiency bottleneck of a self-attention layer, significantly reducing the computational load required for self-attention matrix calculations.
    \item Our experimental results demonstrate that the proposed network outperforms state-of-the-art methods.
\end{itemize}

\section{Related Works}
\subsection{CNN-based SISR}

\begin{figure}
  \centering
    \centering
    \includegraphics[width=\linewidth]{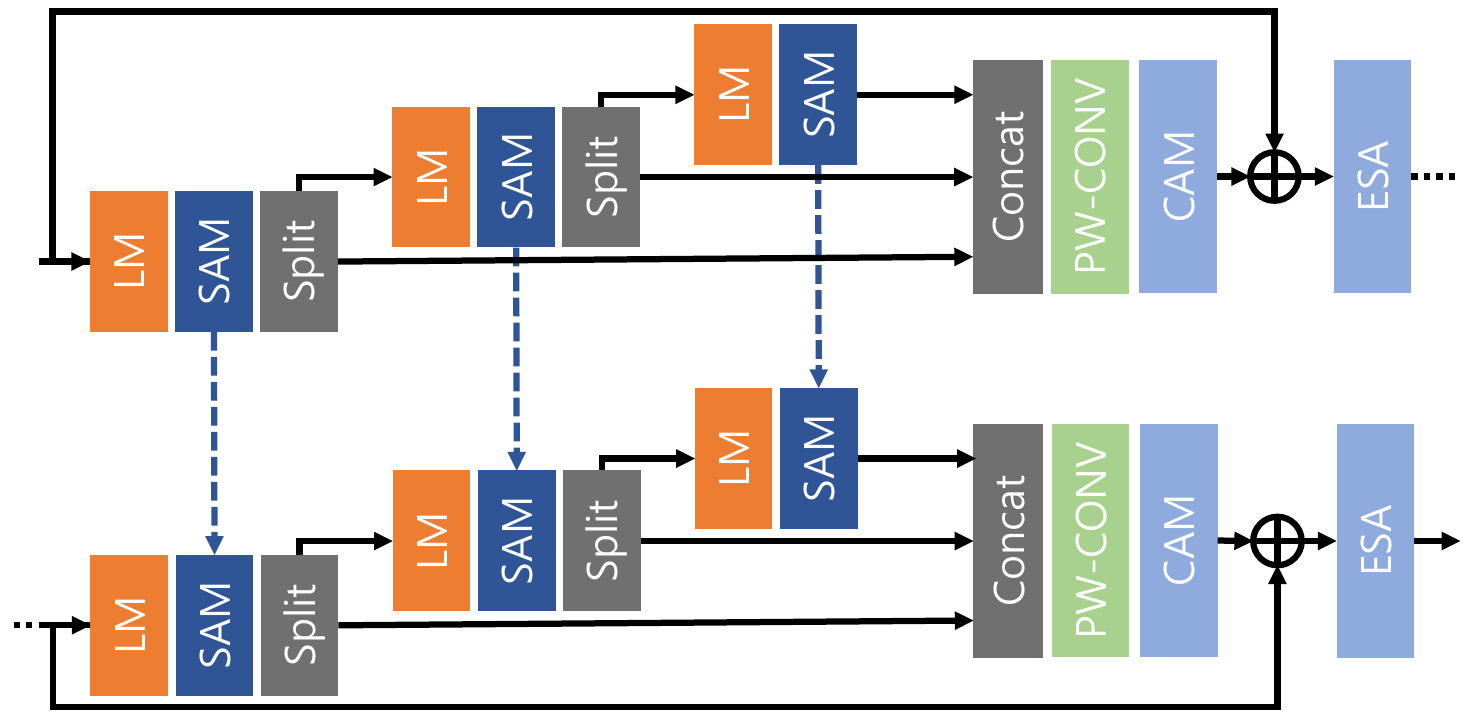}
  \caption{Visualized structure of Information Distillation Blocks (IDBs) and the attention-sharing mechanism. Blue arrows represent the attention-sharing mechanism. PW-CONV denotes pixel-wise convolution.}
  \label{fig:idb}
\end{figure}

\begin{figure*}
  \centering
    \includegraphics[width=0.83\linewidth]{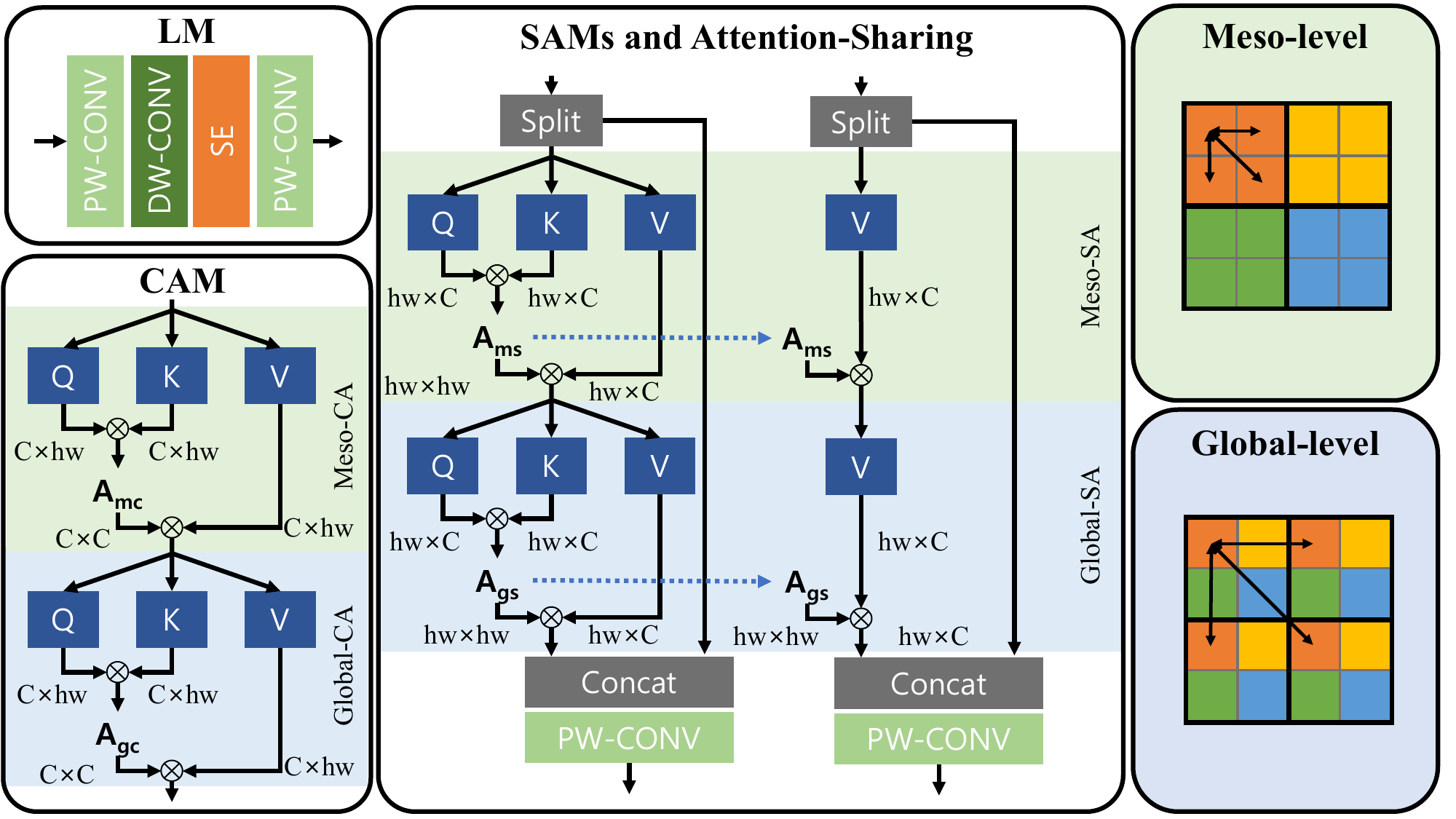}
  \caption{Visualized structure of the Local Module (LM), Spatial Attention Module (SAM), and Channel Attention Module (CAM). The blue arrow represents the attention-sharing mechanism in SAM. By employing the attention-sharing technique, subsequent SAMs bypass the calculation of the spatial attention matrix, which typically accounts for a significant portion of the computational load in self-attention layers. Meso-level self-attention computes attention matrices among pixels within the same partition, whereas global-level self-attention involves pixels from different partitions \cite{wang2023omni}. Utilizing both methods effectively mitigates the limited receptive field issue associated with window-based self-attention. All feedforward networks are omitted for simplicity in visualization.}
  \label{fig:modules}
\end{figure*}

%\subsection{\textbf{Lightweight SR Network}}
There have been many \emph{lightweight networks} for SISR that aimed to find a balance between computational complexity with performance.
Early strategies focused on reusing kernel weights \cite{kim2016deeply,tai2017image,tai2017memnet,li2019feedback}, which effectively reduced the number of parameters but failed to lower computational costs. Additionally, their performance was somewhat limited due to the repetitive use of the same kernels across multiple layers.

To address these shortcomings, researchers also designed efficient and compact network structures. For instance, LatticeNet~\cite{luo2020latticenet} proposed a residual block architecture that exploits multiple potential residual paths. DRSAN~\cite{park2021single} introduced dynamic residual connections that adaptively adjust the residual paths based on the input. 
The information distillation technique~\cite{hui2018fast}, which progressively refines features through the information distillation framework, facilitates efficient computation with low parameters and computational loads. This method has been extensively adopted across various lightweight network designs due to its effectiveness and simplicity \cite{hui2019lightweight,liu2020residual,kong2022residual}.

These lightweight SR networks typically employ shallower architectures with fewer channels to reduce complexity. To compensate for their limited capacity, they enhance feature representation using techniques such as feature reuse, residual connection, and dense connection.

\subsection{Transformer-based SISR}
Recently, there have been significant advancements in vision Transformers \cite{dosovitskiy2020image, liu2021swin, ranftl2021vision}, with several attempts made to apply them to image super-resolution \cite{liang2021swinir,zhang2022efficient,zamir2022restormer,wang2023omni}.
For example,
SwinIR~\cite{liang2021swinir} applied the shifted-window framework~\cite{liu2021swin} for image restoration and demonstrated its effectiveness. ESRT~\cite{lu2022transformer} introduced a new lightweight super-resolution model by combining lightweight CNNs with Transformers.
ELAN~\cite{zhang2022efficient} enhances computational efficiency by sharing the self-attention matrix within a block and the weights of the query and key.
Omni-SR~\cite{wang2023omni} proposed a lightweight Transformer using omni self-attention, which considers spatial and channel self-attention together. SPIN~\cite{zhang2023lightweight} introduced super-pixel clustering into self-attention operations to reduce self-attention computations.
Despite numerous efforts to enhance Transformer efficiency, the structural limitation of stacking self-attention layers continues to hinder their practical application.

\section{\textbf{Proposed Method}}
\label{sec:method}

In this section, we provide a comprehensive overview of our ASID framework. We start by introducing the overall structure of the ASID network, followed by a detailed explanation of modules.

\subsection{Overall Structure}
\label{sec:overall}
Following the previous state-of-the-art (SOTA) method \cite{wang2023omni} as the baseline, we developed a lightweight Transformer structure that utilizes spatial and channel self-attention at both meso- and global-level.
As depicted in \cref{fig:overall}, ASID mainly consists of the convolution layer for shallow feature extraction, repetitive Information Distillation Blocks (IDBs) for deep feature extraction, and Upsampler for image reconstruction. Given the input image as $I_{LR}$, the $3\times3$ convolution layer $H_s$ extracts shallow feature $F_0$ as
\begin{equation}
    F_0 = H_s(I_{LR}).
    \label{eq:shallowfeature}
\end{equation}
This convolutional layer maps the input image from the RGB channel to a multi-channel latent feature dimension.
Next, cascaded IDBs and the $3\times3$ convolution layer $H_d$ extract deep features $F_d$ from shallow features $F_s$ as 
\begin{equation}
    \begin{split}
    & F_1, A = H_{IDB_1}(F_{0}), \\
    & F_i = H_{IDB_i}(F_{i-1}, A),\quad i=2,...N, \\
    & F_d = H_d(F_N),
    \end{split}
    \label{eq:deepfeature}
\end{equation}
where $H_{IDB_i}$ refers to the $i$-th IDB, $A$ means attention matrices for attention-sharing, and $N$ represents the number of IDB. Each IDB consists of multiple self-attention layers, which extract meaningful features by considering long-range dependencies. These stacked IDBs progressively refine input features, with a final convolution layer $H_d$ used to extract deep features.
Then, Upsampler module $H_{up}$ reconstructs high resolution output image $I_{SR}$ from deep features $F_d$ as
\begin{equation}
    I_{SR} = H_{up}(F_0+F_d).
    \label{eq:upsample}
\end{equation}
During this process, the pixel-shuffle layer transforms the latent feature dimension back into the RGB channel and HR space, restoring the output image $I_{SR}$.

\subsection{Information Distillation Block}

Information distillation processes features by hierarchically splitting them during distillation steps. This method preserves essential representations in one part while passing the remaining features to subsequent modules, recognizing that certain feature channels carry more critical information than others~\cite{zhang2018image, wang2021exploring}.
To adapt this approach for Transformer-based SR, we focus on the fundamental structure of the SR Transformer, which can generally be divided into a local feature extraction module and a self-attention calculation module. We design the Information Distillation Block (IDB) for Transformers by aligning the local feature extraction module with convolutional layers for extracting local features and the self-attention modules with sequential calculation units.

As shown in \cref{fig:idb}, IDB refines features in a progressive manner with split and concatenation. IDB is mainly composed of Local Module (LM), Spatial Attention Module (SAM), and Channel Attention Module (CAM). Given the input feature $F_{in}$, $1st$ IDB refines features as
\begin{equation}
    \begin{split}
        & F_1,A_1 = SAM_1(LM_1(F_{in})),\\
        & F_1^{refined},F_1^{coarse} = Split(F_1),\\
        & F_2,A_2 = SAM_2(LM_2(F_1^{coarse})),\\
        & F_2^{refined},F_2^{coarse} = Split(F_2),\\
        & F_3,A_3 = SAM_3(LM_3(F_2^{coarse})),
    \end{split}
    \label{eq:1stidb}
\end{equation}
where $ LM_i$ represents the $i$-th LM, $SAM_i$ means the $i$-th SAM, $A_i$ signifies the $i$-th spatial self-attention matrix, and $ Split$ refers to the channel split. Like the information distillation scheme, the IDB divides features by channels and progressively refines them. LM applies convolution operations to the input features, extracting local features for subsequent self-attention modules. 
Next, SAM enhances features by capturing pixel correlations both within a single window (meso-level) and across different windows (global-level). After progressive feature refinement, refined hierarchical features are aggregated through concatenation and processed by the $1\times1$ convolution layer as

\begin{equation}
    \begin{split}
        & F^{refined} = Conv_{1\times1}(Concat(F_1^{refined},F_2^{refined},F_3),\\
        & F_{out} = ESA(CAM(F^{refined})+F_{in}),
    \end{split}
    \label{eq:1stidb2}
\end{equation}
where $CAM$ represents CAM module,  $Conv_{1\times1}$ refers to pixel-wise convolution, and $ESA$ means enhanced spatial-attention operation \cite{liu2020rfa}. CAM calculates the affinity matrix for intra-window and inter-window pixels across the channel dimension, similar to the operation of SAM. ESA module employs strided convolution and max pooling to achieve a large receptive field and rescale features, thereby enhancing them in a spatial context.

Meanwhile, subsequent IDBs utilize attention-sharing to efficiently perform self-attention layer operations using the spatial self-attention matrices computed by the initial IDB. This can be represented as:
\begin{equation}
    \begin{split}
        & F_1^{refined},F_1^{coarse} = Split(SAM_1(LM_1(F),A_1)),\\
        & F_2^{refined},F_2^{coarse} = Split(SAM_2(LM_2(F_1^{coarse}),A_2)),\\
        & F_3 = SAM_3(LM_3(F_2^{coarse}),A_3),\\
        & F^{refined} = Conv_{1\times1}(Concat(F_1^{refined},F_2^{refined},F_3),\\
        & F_{out} = ESA(CAM(F^{refined})+F).
    \end{split}
    \label{eq:not1stidb}
\end{equation}
These subsequent IDBs skip the self-attention matrix computation by sharing the initial IDB's attention matrix.

\subsubsection{Local Module}
\label{sec:local}
We use LM shown in \cref{fig:modules} to extract local features for self-attention layers. LM is composed of pixel-wise convolution, depth-wise convolution, and a squeeze-and-excitation \cite{hu2018squeeze} module. As shown in \cref{fig:idb}, LM is positioned in front of SAM so that SAM can utilize local information to calculate the spatial correlation matrix.

\subsubsection{Spatial and Channel Attention Module}
\label{sec:sam}
The proposed SAM and CAM employ a window-based self-attention method that processes input features into non-overlapping patches. Both modules operate in two stages, referred to as meso-level and global-level~\cite{wang2023omni}. As shown in \cref{fig:modules}, the meso-level self-attention layer calculates the affinity matrix for pixels within individual patches, while the global-level self-attention layer computes the affinity matrix for pixels across different patches. By incorporating information both within and between patches, the proposed method enhances the network's representational capacity, effectively capturing both local information and long-range dependencies.

Firstly, the input feature $f$ is partitioned into windows for both meso-level and global-level attention. For meso-level, the input feature is partitioned into $P\times P$ non-overlapping patches: $(HW\times C) \rightarrow  ((h\times P)(w\times P)\times C) \rightarrow (hw\times P^2\times C)$, where $P$ represents the partition size of meso-level self-attention, and $\rightarrow$ means reshape operation. For global-level, the input feature is partitioned into $G\times G$ non-overlapping patches: $(HW\times C) \rightarrow  ((G\times h)(G\times w)\times C) \rightarrow (hw\times G^2\times C)$, where $G$ represents the partition size of global-level self-attention.

Then, a linear projection is applied to the input feature to compute the self-attention matrix. Given the input feature $f$, the query $f_{q}$, key $f_{k}$, and value $f_{v}$ are calculated as
\begin{equation}
    f_{q} = Q(f),\quad f_{k} = K(f),\quad f_{v} = V(f),
    \label{eq:sqkv}
\end{equation}
where $Q, K, V$ represent linear projection operation. Next, self-attention matrix $A$ is calculated as
\begin{equation}
    \begin{split}
        & A = SoftMax(f_{q}f_{k}^T), \\
        & f_{out} = FFN(Af_{v}+f),
    \end{split}
    \label{eq:spatialattention}
\end{equation}
where $f_{out}$ represents the output of self-attention layer, and $FFN$ refers to feedforward network.

\begin{table*}[t]
    \centering
    {\small
    \begin{tabular}{clcccccc}
         \multirow{2}{*}{Scale} & \multirow{2}{*}{Methods} &  \multirow{2}{*}{Params} & Set5 & Set14 & B100 & Urban100\\
         & & & PSNR/SSIM & PSNR/SSIM & PSNR/SSIM & PSNR/SSIM\\ \hline \hline
         \multirow{12}{*}{$\times2$} &  \textbf{ASID$^{\dagger}$} &  \textbf{298K} &  \textbf{38.17/0.9613} &  \textbf{33.96/0.9216} &  \textbf{32.31/0.9017} &  \textbf{32.88/0.9353}\\
         & MAFFSRN & 402K & 37.97/0.9603 & 33.49/0.9170 & 32.14/0.8994 & 31.96/0.9268\\
         & SPIN$^{\dagger}$ & 497K & 38.20/0.9615 & 33.90/0.9215 & 32.31/0.9015 & 32.79/0.9340\\
         & RFDN  & 534K & 38.05/0.9606 & 33.68/0.9184 & 32.16/0.8994 & 32.12/0.9278\\
         & ELAN$^{\dagger}$ & 582K & 38.17/0.9611 & 33.94/0.9207 & 32.30/0.9012 & 32.76/0.9340\\
         & ESRT$^{\dagger}$ & 677K & 38.03/0.9600 & 33.75/0.9184 & 32.25/0.9001 & 32.58/0.9318 & \\
         & DRSAN & 690K & 38.11/0.9609 & 33.64/0.9185 & 32.21/0.9005 & 32.35/0.9304 & \\
         & IMDN & 694K & 38.00/0.9605 & 33.63/0.9177 & 32.19/0.8996 & 32.17/0.9283 & \\
         & \textbf{ASID-D8}$^{\dagger}$ &  \textbf{732K} &  \textbf{38.32/0.9618} &  \textbf{34.24/0.9232} &  \textbf{32.40/0.9028} &  \textbf{33.35/0.9387}\\
         & LatticeNet & 756K & 38.15/0.9610 & 33.78/0.9193 & 32.25/0.9005 & 32.43/0.9302 & \\
         & Omni-SR$^{\dagger}$ & 772K & 38.22/0.9613 & 33.98/0.9210 & 32.36/0.9020 & 33.05/0.9363 & \\
          & SwinIR$^{\dagger}$ & 878K & 38.14/0.9611 & 33.86/0.9206 & 32.31/0.9012 & 32.76/0.9340 & \\
         \hdashline
         \multirow{11}{*}{$\times3$} & \textbf{ASID$^{\dagger}$} & \textbf{304K} & \textbf{34.58/0.9290} & \textbf{30.53/0.8465} & \textbf{29.23/0.8096} & \textbf{28.68/0.8633}\\
         & RFDN  & 541K & 34.41/0.9273 & 30.34/0.8420 & 29.09/0.8050 & 28.21/0.8525\\
         & SPIN$^{\dagger}$ & 569K & 34.65/0.9293 & 30.57/0.8464 & 29.23/0.8089 & 28.71/0.8627\\
         & ELAN$^{\dagger}$ &  590K & 34.64/0.9288 & 30.55/0.8463 & 29.21/0.8081 & 28.69/0.8624\\
         & IMDN & 703K & 34.36/0.9270 & 30.32/0.8417 & 29.09/0.8046 & 28.17/0.8519 & \\
         & \textbf{ASID-D8}$^{\dagger}$ & \textbf{739K} & \textbf{34.84/0.9307} & \textbf{30.66/0.8491} & \textbf{29.32/0.8119} & \textbf{29.08/0.8706}\\
         & DRSAN & 740K & 34.50/0.9278 & 30.39/0.8437 & 29.13/0.8065 & 28.35/0.8566 & \\
         & LatticeNet & 765K & 34.53/0.9281 & 30.39/0.8424 & 29.15/0.8059 & 28.33/0.8538 & \\
         & ESRT$^{\dagger}$ & 770K & 34.42/0.9268 & 30.43/0.8433 & 29.15/0.8063 & 28.46/0.8574 & \\
         & Omni-SR$^{\dagger}$ & 780K & 34.70/0.9294 & 30.57/0.8469 & 29.28/0.8094 & 28.84/0.8656 & \\
         & MAFFSRN & 807K & 34.45/0.9277 & 30.40/0.8432 & 29.13/0.8061 & 28.26/0.8552\\
         & SwinIR$^{\dagger}$ & 886K & 34.62/0.9289 & 30.54/0.8463 & 29.20/0.8082 & 28.66/0.8624 & \\
         \hdashline
         \multirow{12}{*}{$\times4$} & \textbf{ASID$^{\dagger}$} & \textbf{313K} & \textbf{32.39/0.8964} & \textbf{28.80/0.7865} & \textbf{27.70/0.7418} & \textbf{26.48/0.7978}\\
         & RFDN  & 550K & 32.24/0.8952 & 28.61/0.7819 & 27.57/0.7360 & 26.11/0.7858\\
         & SPIN$^{\dagger}$ & 555K & 32.48/0.8983 & 28.80/0.7862 & 27.70/0.7415 & 26.55/0.7998\\
         & ELAN$^{\dagger}$ &  601K & 32.43/0.8975 & 28.78/0.7858 & 27.69/0.7406 & 26.54/0.7982 \\
         & IMDN & 715K & 32.21/0.8948 & 28.58/0.7811 & 27.56/0.7353 & 26.04/0.7838 & \\
         & DRSAN & 730K & 32.30/0.8954 & 28.66/0.7838 & 27.61/0.7381 & 26.26/0.7920 & \\
         & \textbf{ASID-D8}$^{\dagger}$ & \textbf{748K} & \textbf{32.57/0.8990} & \textbf{28.89/0.7898} & \textbf{27.78/0.7449} & \textbf{26.89/0.8096}\\
         & ESRT$^{\dagger}$ & 751K & 32.19/0.8947 & 28.69/0.7833 & 27.69/0.7379 & 26.39/0.7962 & \\
         & LatticeNet & 777K & 32.30/0.8962 & 28.68/0.7830 & 27.62/0.7367 & 26.25/0.7873 & \\
         & Omni-SR$^{\dagger}$ & 792K & 32.49/0.8988 & 28.78/0.7859 & 27.71/0.7415 & 26.64/0.8018 & \\
         & MAFFSRN & 830K & 32.20/0.8953 & 26.62/0.7822 & 27.59/0.7370 & 26.16/0.7887\\
         & SwinIR$^{\dagger}$ & 897K & 32.44/0.8976 & 28.77/0.7858 & 27.69/0.7406 & 26.47/0.7980 & \\ \hline
    \end{tabular}
    }
    \caption{Quantitative comparison of previous CNN-based lightweight SR models and the proposed method. The $\dagger$ symbol represents a Transformer-based SR method, and \textbf{bold} highlights the proposed method.}
    \label{tab:modelstudy}
\end{table*}

\begin{table}[t]
    \centering
    %\hspace*{-0.75cm}
    {%\small
    \begin{tabular}{ccccccc}
         Scale &  Methods & Params & FLOPs & PSNR \\ \hline \hline
         \multirow{6}{*}{$\times4$} & \textbf{ASID} & \textbf{313K} & \textbf{17.2G} & \textbf{28.80} \\
          & ELAN & 601K & 43.2G & 28.78 \\
          & IMDN & 715K & 40.9G & 28.58 \\
          & LatticeNet & 777K & 43.6G & 28.68 \\
          & SwinIR & 897K & 49.6G & 28.77 \\
          & \textbf{ASID-D8} & \textbf{748K} & \textbf{40.7G} & \textbf{28.89} \\
         \hline
          %& ASID-D8 & 748k & 40.7G & 28.89 \\ \hline
    \end{tabular}
    }
    \caption{Comparisons of the computational cost of lightweight SR methods on the Set14 ($\times4$) dataset. FLOPs are evaluated on a 720p ($1280\times720$) output image. \textbf{Bold} indicates the proposed method.}
    \label{tab:computation}
\end{table}

\subsubsection{Attention-Sharing and Channel-Split}
The computational burden of self-attention is a key inefficiency in SR Transformers, yet stacking spatial self-attention layers is essential for performance. To address this, we propose attention-sharing and channel-split techniques.
As shown in \cref{fig:modules}, attention-sharing enables the sharing of attention matrices between self-attention layers, eliminating the need for spatial self-attention operations. This allows layers that share attention matrices to skip the computation of self-attention matrices, reducing the parameters required for affinity matrix calculation. Channel-split restricts the number of channels involved in spatial attention operations, which helps decrease both the computational load and the number of parameters. The combination of attention-sharing and channel-split reduces the complexity of self-attention, enabling the stacking of more layers with fewer parameters.

\begin{figure*}
  \centering
    \centering
    \includegraphics[width=\linewidth]{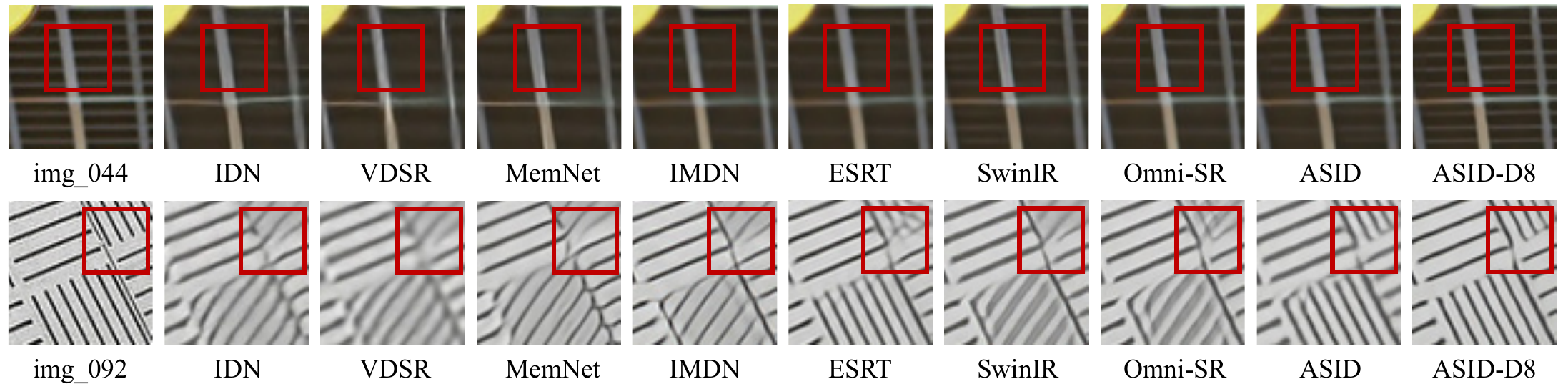}
  \caption{Qualitative Comparison of previous CNN-based and Transformer-based SR methods on the Urban100 ($\times4$) dataset. Note that ASID accurately restores images while using an extremely small number of model parameters.}
  \label{fig:viscom}
\end{figure*}

\section{Experimental Results}
\subsection{Settings}

Following previous methods, the DIV2K dataset~\cite{agustsson2017ntire} is selected for training, which contains 800 HR images.
During training, 64$\times$64-sized RGB patches are utilized, with random flips and rotations applied for data augmentation.
%We use 64$\times$64-sized patches in RGB, along with their random flips and rotations, as the training dataset. 
For evaluation, we use well-known benchmark datasets, including Set5~\cite{bevilacqua2012low}, Set14~\cite{zeyde2010single}, B100~\cite{martin2001database}, and Urban100~\cite{huang2015single}.
The performance of the network is evaluated using PSNR and SSIM on the Y channel.
For training, we use the ADAM optimizer with a mini-batch size of 16. All models in quantitative comparisons are trained for $1000$ epochs, while the models used in ablation studies are trained for $200$ epochs. The initial learning rate is set to $5\times10^{-4}$ and is halved every $250$ epochs. 

We set the number of IDBs in ASID to 3 and the number of channels to 48. Also, we set the number of calculation units in each IDB to 3, which include LM, SAM, and channel-split layers as shown in \cref{fig:idb}.
In the channel-split layers within or following SAM, we allocate 12 channels to the refined features, while the remaining channels are designated as coarse features and passed on to the next calculation unit. For the self-attention layers, we set the partition size for meso-level attention to 8 and the grid size for global-level attention to 8.

\subsection{Comparisons with SOTA Methods}
In this section, we compare our ASID with state-of-the-art (SOTA) lightweight CNN-based and Transformer-based SR methods to demonstrate the proposed method's effectiveness. We report quantitative and qualitative comparisons with previous SR methods, including DRSAN \cite{park2021single}, ECBSR \cite{zhang2021edge}, ELAN \cite{zhang2022efficient}, ESRT \cite{lu2022transformer}, FALSR \cite{chu2019fast}, IDN \cite{hui2018fast}, IMDN \cite{hui2019lightweight}, LapSRN \cite{lai2018fast}, LatticeNet \cite{luo2020latticenet}, MAFFSRN \cite{muqeet2020multi}, MemNet \cite{tai2017memnet}, Omni-SR \cite{wang2023omni}, RFDN \cite{liu2020residual}, SPIN \cite{zhang2023lightweight}, SwinIR \cite{liang2021swinir}, and VDSR \cite{Kim_2016_VDSR}.

We provide quantitative comparisons between proposed methods and previous lightweight SR methods in \cref{fig:perfcomp,,tab:modelstudy}.
Despite having fewer model parameters, our method significantly outperforms CNN-based SR methods, as shown in \cref{tab:modelstudy}.
%\cref{tab:modelstudy} compares ASID with previous lightweight SR methods. Despite having fewer model parameters, our method significantly outperforms CNN-based SR methods.
These results suggest that while CNN-based SR methods mainly process local features, ASID utilizes self-attention to capture long-range dependencies. This leads to a broader receptive field, improving image reconstruction accuracy. Moreover, ASID achieves performance comparable to previous Transformer-based SR methods, using less than 60$\,\%$ of their model parameters.
This demonstrates that ASID effectively preserves essential information about long-range dependencies, even with lightweight strategies, maintaining performance on par with more complex methods. We further introduce the ASID-D8 model as an additional option, providing an alternative trade-off between performance and complexity. 
As shown in \cref{tab:computation}, our approach not only reduces model parameters but also significantly lowers computational costs compared to prior methods. In \cref{fig:viscom}, we compare zoomed-in results from the Urban100 ($\times4$) dataset, highlighting the superior quality of ASID over previous lightweight methods. More qualitative comparisons and experiments on RealSR dataset~\cite{cai2019toward} are provided in the supplementary material.

\subsection{Ablation Studies}

\begin{table}[t]
    \centering
    {%\small
    \begin{tabular}{ccccccc}
         Methods &  Params & FLOPs & PSNR \\ \hline \hline
         Baseline & \textbf{288K} & \textbf{62.323G} & 32.43   \\
         + ID & 401K & 87.344G & \textbf{32.70}   \\
         + ID \& AS & 365K & 79.119G & 32.62  \\
         + ID \& CS & 316K & 67.836G & 32.53  \\
         + ID \& AS \& CS & 298K & 63.592G & 32.53  \\
         \hline
    \end{tabular}
    }
    \caption{Ablation Studies on the proposed ASID network structure. All results are evaluated on the Urban100 ($\times2$) dataset. ID refers to the Information Distillation framework, AS denotes attention-sharing, and CS represents channel-split.}
    \label{tab:ablation1}
\end{table}

\subsubsection{Information Distillation Block }
We propose lightweight network approaches for ASID, including information distillation structure, attention-sharing, and channel-split. To investigate the effectiveness of these proposed methods, we compare the network performance with and without these elements. For the comparison, we introduce a baseline structure consisting of serially connected LM, SAM, and CAM. The results of the experiments are summarized in \cref{tab:ablation1}.

First, we observe that the baseline performance is the lowest among all configurations. Subsequently, applying the information distillation scheme to the baseline results in an increase in performance, which is primarily attributed to the doubling of the number of self-attention layers. However, this increase in self-attention layers also leads to a 40$\,\%$ rise in both model parameters and computational cost.

Next, we observe that implementing attention-sharing reduces the model parameters and computational load by $10\,\%$ each. With the channel-split method, we achieve a $20\,\%$ reduction in parameters and computational cost individually. By integrating both methods, the network achieves enhanced performance while maintaining a complexity level that is nearly identical to the baseline. This underscores the effectiveness of these lightweight strategies in enhancing the network's overall efficiency.

\subsubsection{Attention-Sharing }
\begin{figure}
  \centering
    \centering
    \includegraphics[width=\linewidth]{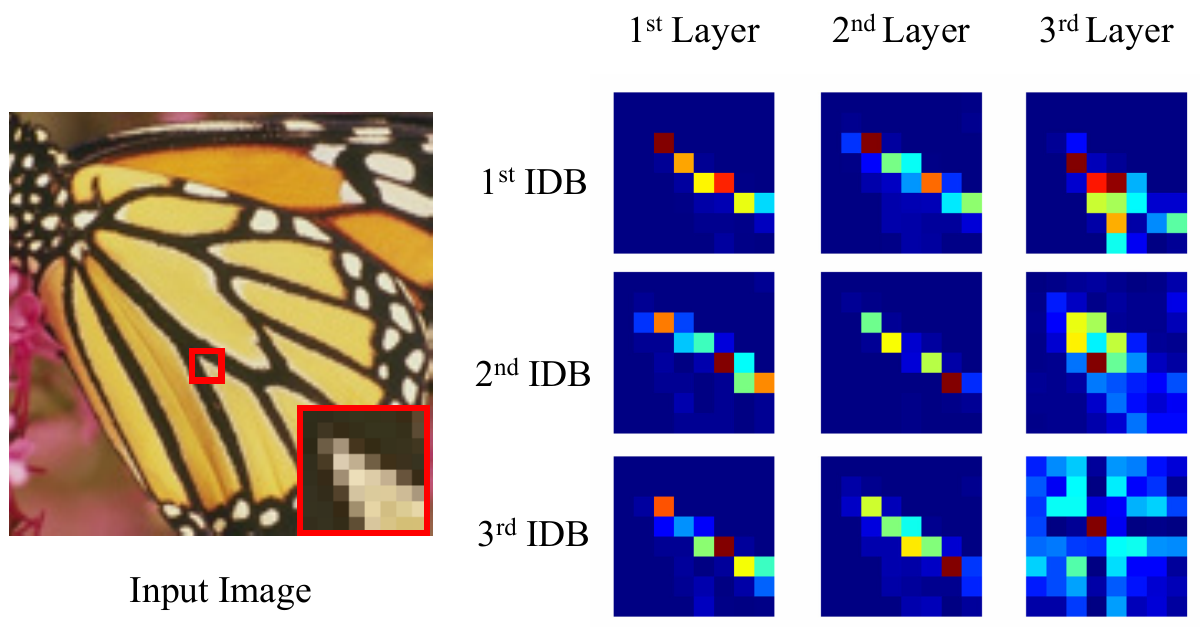}
  \caption{Visualized comparison of attention matrices. The attention matrices are collected from models without attention-sharing, as described in \cref{tab:ablation1}. The visualization depicts the meso-level spatial correlation between the center point and the pixels within the same window.}
  \label{fig:attcomp}
\end{figure}

To investigate suitable attention-sharing methods for our proposed structure, we visualize the spatial attention from an ablation model without attention-sharing in \cref{tab:ablation1}, specifically focusing on the spatial correlations around the center point for easier visualization. As \cref{fig:attcomp} shows, even within the same block, the use of spatial attention matrices varies with depth. Therefore, we chose the sharing of spatial attention matrices across blocks of the same depth rather than within the same block. This approach enables the IDB to enhance features by considering various spatial correlations.

\begin{table}[t]
    \centering
    {%\small
    \begin{tabular}{ccccccccc}
         Scale & Methods &  Params & FLOPs & PSNR\\
         \hline \hline
          \multirow{2}{*}{$\times2$} & IntraGroup & 305K & 65.317G & 32.49\\
          & InterGroup & 298K & 63.592G & 32.53\\
         \hline
    \end{tabular}
    }
    \caption{Ablation studies on attention-sharing methods. IntraGroup refers to attention-sharing between adjacent layers within the same block, while InterGroup denotes attention-sharing across blocks, as proposed in our method. PSNR results are evaluated on the Urban100 ($\times2$) dataset.}
    \label{tab:ablationAS}
\end{table}

Next, we evaluate the effectiveness of the proposed method by comparing two candidate methods for attention-sharing. \cref{tab:ablationAS} shows two different attention matrix-sharing methods: IntraGroup and InterGroup. IntraGroup shares attention matrices within building blocks, while InterGroup shares attention matrices between building blocks. The major difference between the two methods is that IntraGroup enforces adjacent self-attention layers to use the same attention matrix, whereas InterGroup allows adjacent self-attention layers to use different attention matrices. As shown in \cref{tab:ablationAS}, our proposed method shows better results with fewer parameters and FLOPs. This suggests that allowing adjacent self-attention layers to capture diverse pixel correlations improves feature representation.

\subsubsection{Network Depth} 
\begin{figure}
  \centering
    \centering
    \includegraphics[width=0.9\linewidth]{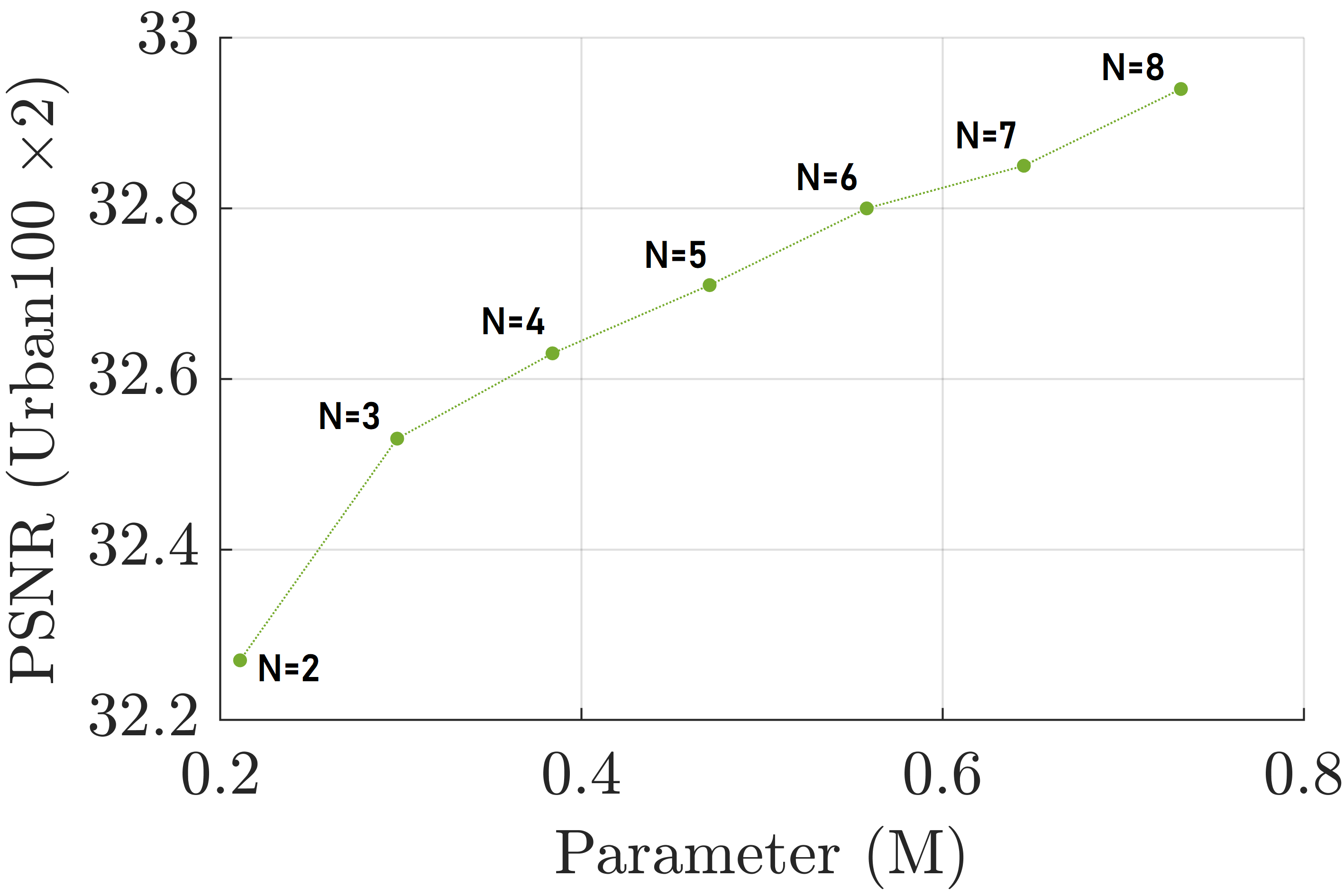}
  \caption{Ablation studies on network depth. Results are evaluated on the Urban100 ($\times2$) dataset. N indicates the number of IDBs implemented in the network.}
  \label{fig:length}
\end{figure}

We examine the effect of network depth on performance by adjusting the number of IDBs $N$ from 2 to 8. As shown in \cref{fig:length}, the network's performance consistently improves with an increasing number of blocks, even though correlation matrix computations are omitted due to attention-sharing from the second IDB onward. This demonstrates that the proposed attention-sharing method effectively supports both shallow and deeper network architectures. Considering the trade-off between performance improvements and network complexity, we determine the optimal depth for ASID to be $N=3$, where performance gains start to diminish, allowing us to retain minimal complexity within the lightweight ASID framework.

\section{Conclusion}
We propose the attention-sharing information distillation (ASID) network, a novel lightweight Transformer-based SR method that delivers competitive performance compared to existing lightweight SR methods while utilizing significantly fewer model parameters.
ASID employs an information distillation structure specifically adapted for Transformers, enabling the efficient stacking of multiple self-attention layers with low complexity. Additionally, ASID incorporates attention-sharing and channel-split techniques to significantly reduce the computational overhead typically associated with self-attention operations.
Experimental results demonstrate that ASID effectively balances model complexity with performance, surpassing previous lightweight SR methods.

\section{Acknowledgments}
This work was supported in part by Samsung Electronics Co., Ltd., and in part by Institute of Information $\&$ Communications Technology Planning $\&$ Evaluation (IITP) grant funded by the Korea government (MSIT) [NO.RS-2021-II211343, Artificial Intelligence Graduate School Program (Seoul National University)]

\bigskip

\bibliography{aaai25}

\end{document}

\maketitle

\section{Ablation Studies on Channel-Split}

\begin{figure}
  \centering
    \centering
    \includegraphics[width=\linewidth]{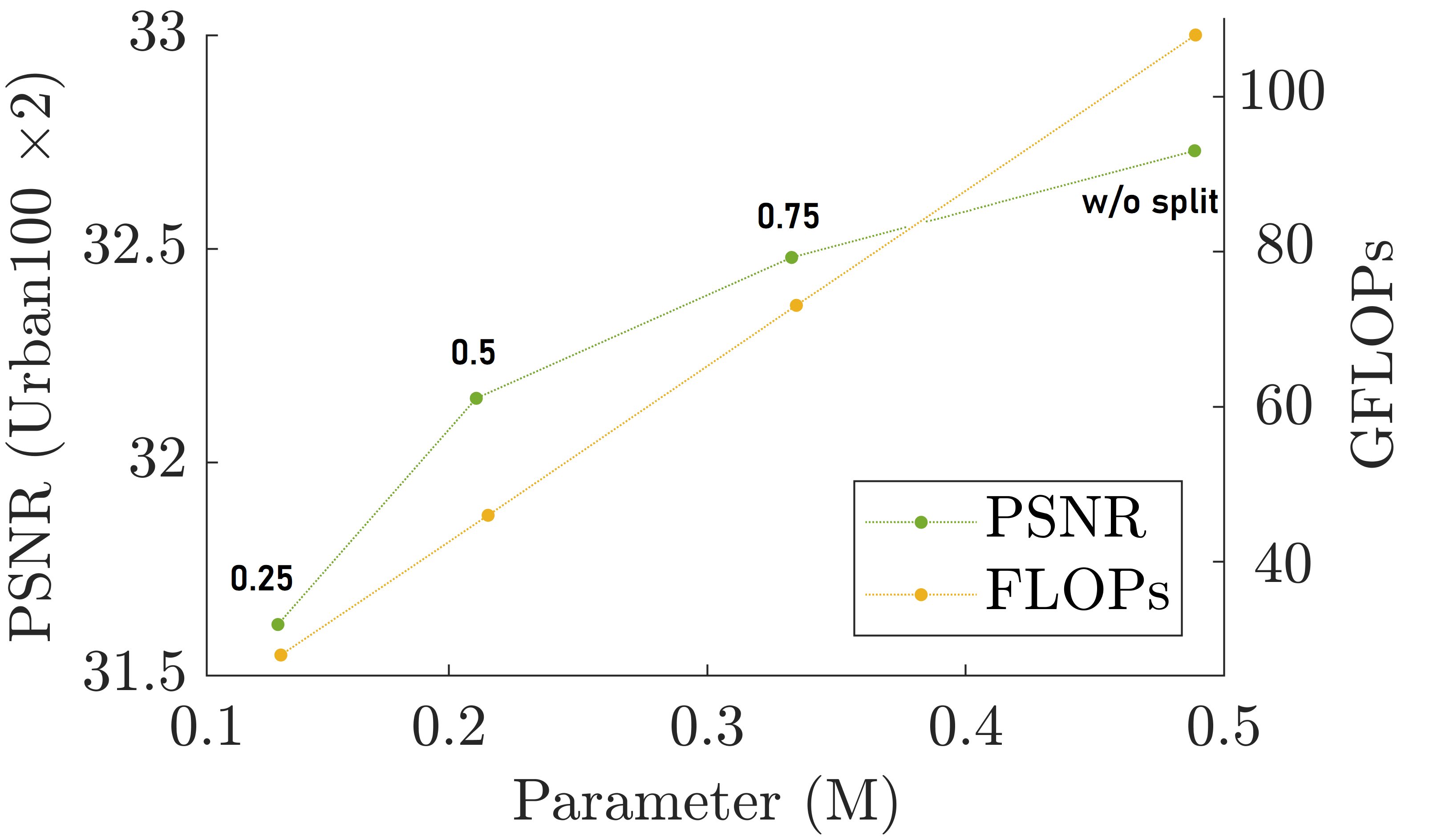}
  \caption{Ablation studies on channel-split methods. Results are evaluated on the Urban100 ($\times2$) dataset. The ratio represents how many channels from the entire feature are fed into self-attention layers.}
  \label{fig:cs}
\end{figure}

For ASID, we limit the number of feature channels fed into the spatial self-attention layer to reduce the model parameters and computational load. To evaluate the impact of channel-split on reconstruction performance, we analyze how changes in the number of channels fed into the self-attention layer influence overall performance. \cref{fig:cs} illustrates the relationship between network performance and the ratio of channels processed by the self-attention module. Omni-SR \cite{wang2023omni} with three blocks is selected as our baseline for comparison. As demonstrated in \cref{fig:cs}, the computational load decreases proportionally with the ratio of channels. Meanwhile, the network's performance deteriorates at an increasing rate as a larger proportion of channels bypasses the self-attention layer. We set the channel-split ratio at 0.75 for the first IDB to balance the performance and network complexity. 

\section{Real Image Super-Resolution}

\begin{table}[t]
    \centering
    %\hspace*{-0.75cm}
    \caption{Comparisons of SISR methods on RealSR dataset. \textbf{Bold} represents the proposed method.}
    {%\small
    \begin{tabular}{cccc}
         Scale &  Methods & PSNR & SSIM \\ \hline \hline
         \multirow{4}{*}{$\times3$} & IMDN & 30.29 & 0.857 \\
          & LP-KPN & 30.60 & 0.863 \\
          & ESRT & 30.38 & 0.857 \\
          & \textbf{ASID} & \textbf{30.91} & \textbf{0.868} \\ \hline
          \multirow{4}{*}{$\times4$} & IMDN & 28.68 & 0.815 \\
          & LP-KPN & 28.65 & 0.820 \\
          & ESRT & 28.78 & 0.815 \\
          & \textbf{ASID} & \textbf{29.27} & \textbf{0.829} \\ \hline
          %& ASID-D8 & 748k & 40.7G & 28.89 \\ \hline
    \end{tabular}
    }
    \label{tab:realsr}
\end{table}

Additionally, to evaluate the network's performance under unknown degradation rather than bicubic degradation, we perform training and evaluation on the RealSR (V3) dataset~\cite{cai2019toward}. During training, ASID models pre-trained on bicubic degradation are fine-tuned using the RealSR dataset. For the evaluation, we compare lightweight SR methods ~\cite{hui2019lightweight,cai2019toward,lu2022transformer} applied to real-world super-resolution tasks. As shown in the table, despite not being specifically designed for the RealSR dataset, the proposed ASID consistently demonstrated strong performance across various metrics. These results highlight the potential of the proposed method to handle degradations beyond bicubic degradation and its effectiveness for real-world images.

\section{Discussions}
In this section, we discuss the advantages of the proposed ASID framework and its differences from previous methods.

\subsection{ASID Framework}
The benefits of the ASID framework can be summarized as follows.
First, the information distillation scheme directly addresses the efficiency bottleneck of correlation matrix calculation, a common limitation in self-attention layers. The proposed structure reduces the number of channels involved in self-attention operations as the network deepens, alleviating computational inefficiencies. This enables ASID to stack multiple self-attention layers with fewer parameters and less computational cost compared to other Transformer-based SR methods.
Second, attention-sharing significantly reduces the computational cost of self-attention layer operations. The spatial self-attention matrices, which define pixel relationships, are calculated in the 1st IDB and shared across subsequent IDBs, eliminating a substantial portion of the self-attention operations. This strategy allows ASID to further reduce parameters and computational load while stacking self-attention layers.
By combining these two strategies, ASID achieves high performance with substantially fewer operations and parameters than previous SR methods.

\subsection{Comparison with Previous SR Methods} 
First, we review previous methods similar to ASID, both CNN-based and Transformer-based SR approaches, and highlight the key differences between them.

\subsubsection{Information Distillation Methods } The network design technique known as information distillation \cite{hui2018fast,hui2019lightweight,liu2020residual,kong2022residual} is well-known for its efficient computation with minimal parameters, making it a popular choice in various CNN-based lightweight SR methods. Unlike traditional information distillation methods, which primarily enhance the efficiency of CNNs for handling local information, the Transformer-based ASID focuses on capturing long-range pixel dependencies through self-attention operations.
Previous information distillation methods employ a series of calculation units that gradually reduce the channels involved in feature processing, thereby improving the efficiency of local feature extraction. In contrast, ASID integrates channel-split directly within the self-attention layers, rather than in separate modules dedicated to local feature extraction. This design maintains the Transformer’s strength in capturing long-range dependencies.
Moreover, we introduce an attention-sharing technique to address the significant computational bottleneck of self-attention layers in ASID. While previous information distillation methods use relatively lightweight calculation units due to their focus on local features, ASID faces an increased computational load as more self-attention layers are added. By sharing the self-attention matrix across layers, ASID reduces this computational burden without compromising network performance.

\subsubsection{Transformer-based SR Methods } Significant research has been conducted to enhance the practicality of Transformer-based SR methods. Zhang {\etal} \shortcite{zhang2022efficient} proposed ELAN, which reduces network parameters and computational load by sharing the attention matrix. The primary difference between ELAN and ASID lies in their attention-sharing mechanisms and overall network design. ELAN shares a spatial attention matrix within the self-attention layers of the same building block. In contrast, our method computes all attention matrices in the first block, with subsequent blocks reusing these pre-computed matrices.
This difference in sharing strategy enables ASID to allow adjacent layers to use different attention matrices, thereby capturing a more diverse set of pixel correlations compared to ELAN, where adjacent self-attention layers share the same spatial attention matrix. For a detailed comparison of these two attention-sharing strategies, please refer to the ablation study.
Furthermore, the proposed information distillation structure in ASID enables the construction of a network with fewer parameters than ELAN, improving efficiency while enhancing its ability to capture and utilize complex image details in SR tasks.

Next, Wang {\etal} \shortcite{wang2023omni} proposed Omni Self-Attention (OSA), which considers pixel correlations at both meso-level and global-level by combining spatial and channel self-attention. While ASID shares a similar architecture with Omni-SR, a key difference lies in how attention is processed within the network. Omni-SR uses OSA to aggregate spatial and channel attention. In contrast, based on ablation studies from previous work~\cite{wang2023omni}, we determined that spatial self-attention contributes more significantly to Transformer performance than channel self-attention. Therefore, ASID separates spatial and channel attention, stacking multiple spatial self-attention layers.
This architectural distinction means that Omni-SR calculates meso-level omni self-attention before global-level omni self-attention. In contrast, ASID computes spatial self-attention at both meso- and global-levels across the stacked spatial self-attention layers, with channel self-attention performed at the end of the module. Additionally, while Omni-SR relies solely on window-based self-attention to address the efficiency bottleneck, ASID achieves greater efficiency by incorporating the attention-sharing information distillation scheme. This not only mitigates computational challenges more effectively but also improves the overall performance of the lightweight Transformer-based SR network.

\begin{figure*}
  \centering
    \centering
    \includegraphics[width=0.75\linewidth]{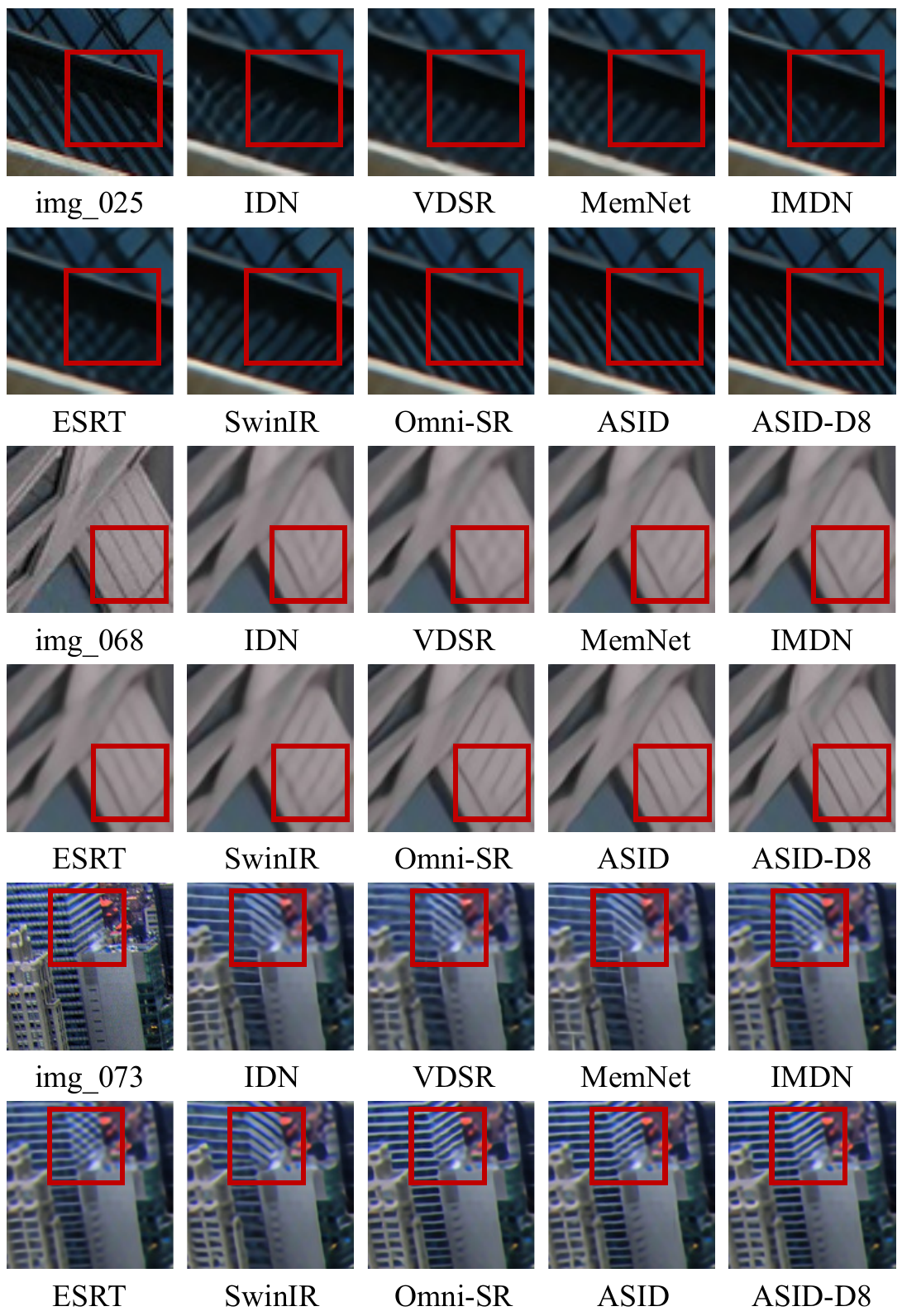}
  \caption{Qualitative Comparison of previous CNN-based and Transformer-based SR methods on Urban100 ($\times4$) dataset. Note that ASID can restore accurate images with extremely few model parameters.}
  \label{fig:viscom}
\end{figure*}

\section{Qualitative Comparison } In \cref{fig:viscom}, we compare zoomed-in results of previous lightweight SR methods and ASID on the Urban100 ($\times4$) dataset. As shown in \cref{fig:viscom}, ASID reconstructs SR images with precise detail while using the fewest model parameters, totaling only 313K.
Notably, ASID produces fewer artifacts despite having significantly fewer parameters than other models. Furthermore, ASID restores SR images with quality comparable to Transformer-based methods with more than twice the parameters. This performance can be attributed to the proposed information distillation structure, which enables efficient stacking of self-attention layers and effectively expands the receptive field. This improvement is achieved without a substantial increase in parameters, underscoring the efficiency of ASID’s lightweight design.

\bibliography{supplement}